\documentclass[conference]{IEEEtran}
\IEEEoverridecommandlockouts
\usepackage{cite}
\usepackage{amsmath,amssymb,amsfonts}
\usepackage{algorithmic}
\usepackage{graphicx}
\usepackage{subcaption}
\usepackage{stfloats} 
\usepackage{multirow} 
\usepackage{booktabs}

\usepackage{textcomp}
\usepackage{xcolor}
\usepackage{hyperref}
\usepackage[utf8]{inputenc}
\usepackage{scrextend}
\usepackage{siunitx}

\usepackage[nomargin,inline,draft]{fixme}
\fxsetup{theme=color,mode=multiuser}
\FXRegisterAuthor{j}{jla}{\color{purple}JLA}
\FXRegisterAuthor{r}{rdm}{\color{teal}RdM}

\def\BibTeX{{\rm B\kern-.05em{\sc i\kern-.025em b}\kern-.08em
    T\kern-.1667em\lower.7ex\hbox{E}\kern-.125emX}}

\begin{document}

\title{A Spatio-Temporal Spot-Forecasting Framework for Urban Traffic Prediction}

\author{\IEEEauthorblockN{Rodrigo de Medrano, José L. Aznarte}
\IEEEauthorblockA{\textit{Artificial Intelligence Department} \\
\textit{Universidad Nacional de Educación a Distancia --- UNED}\\
Madrid, Spain \\
\texttt{jlaznarte@dia.uned.es}}
}

\maketitle

\begin{abstract}
  Spatio-temporal forecasting is an open research field whose interest is
  growing exponentially. In this work we focus on creating a complex deep neural
  framework for spatio-temporal traffic forecasting with comparatively very good
  performance and that shows to be adaptable over several spatio-temporal
  conditions while remaining easy to understand and interpret. Our proposal is
  based on an interpretable attention-based neural network in which several
  modules are combined in order to capture key spatio-temporal time series
  components. Through extensive experimentation, we show how the results of our
  approach are stable and better than those of other state-of-the-art
  alternatives. 
\end{abstract}

\begin{IEEEkeywords}
  deep learning, neural networks, spatio-temporal series, traffic forecasting
\end{IEEEkeywords}

\section{Introduction}
\label{S1}


Spatio-temporal forecasting is playing a key role in our efforts to understand
and model environmental, operational and social processes of all kinds and their
interrelations all over the globe. From climate science and transportation
systems to finances and economic, there are plenty of fields in which time and
space might constitute two entangled dimensions of data, with one affecting the
other and thus both being relevant for prediction. In this context, there is an
increasing trend to develop and improve methodologies for gathering and using
vast amounts of spatio-temporal data over the last years. Tailored to extract
useable knowledge from these big data repositories, there are plenty of
proposals trying to facilitate a shared understanding of the multiple
relationships between the physical and natural environments and society (being
the UE's projects Digital Earth \cite{noauthor_digital_nodate} or Galileo
\cite{noauthor_galileo_nodate} two salient examples). By contributing in this
direction, it is possible to enrich a great deal of services in many ways and
gain a better understanding of our world. While machine learning has been widely
used for spatio-temporal forecasting in the last decade, there is still room for
improvement in our understanding of the models and in their applications.

Specifically, when using neural networks (NN) for regression tasks it is highly
desirable that these intelligent systems are capable of adapting to a wide range
of circumstances within the framework in which they have been trained. As this
ability depends on the data and problem in which the NN is being applied, every
field might present different aspects in which it could be beneficial.

In the concrete case of spatio-temporal forecasting, the prediction depends
fundamentally on two dimensions: the time horizon and the spatial zone in which
the NN is being trained. Thus, traditionally NN are trained and evaluated over
some fixed spatial and temporal conditions, restricting the contexts in which
they can be applied, making them less suitable to deal with atypical inputs and
limiting the knowledge about its general behaviour. While creating a system that
can infer future properties of the series with a single training is out of the
scope with actual techniques, it is important to evaluate algorithms over
different spatio-temporal scenarios as every methodology usually presents
dissimilar behaviours in distinct situations. Thus, even if a fixed application
is intended, exploring the adaptability to different circumstances of an
algorithm might be positive.

On the contrary, we propose to characterize spatio-temporal frameworks via a
complete and comprehensive experimentation and evaluation over both dimensions.
This evaluation methodology, which has been named for convenience as
spot-forecasting (in analogy with the economic term \textit{Spot Market}),
explores the adaptability of neural systems to any spatio-temporal input for a
specific series. Its name refers to the property of these models to predict at
any moment in which the forecast is needed. Given some forecasting conditions as
number of input-output timesteps and spatial points, the idea is to train and
evaluate the network with any possible temporal sequence from the series for a
wide range of spatial allocations of different nature. For example, instead of
making 24 hours prediction starting at 00.00 every day for each point, making 24
hours predictions whose start can be any possible hour of the day. Even if a
system will work under a rigid scenario, this strategy lets us gain a wider
insight of the model, facilitates its application to other spatio-temporal
conditions (directly or via transfer learning), makes a more robust model to
unusual inputs and works as a data-augmentation technique due to the increase of
training population (in the previous example, from one training sample per day
to 24 training samples per day).

Pointing in this direction, through this work we propose a novel Neural Network
called CRANN (from Convo-Recurrent Attentional Neural Network) that is evaluated
for several spatio-temporal conditions and compared with some of the state of
the art methods. The model presented in this paper is built on the idea of the
classical time series decomposition, which attempts to separately model the
available knowledge about the underlying unknown generator process. This
generator process is usually considered to be composed of several terms like
seasonalities, trend, inertia and spatial relations, plus noise. Thus, our model
is defined like a composition of several modules that exploit different neural
architectures in order to separately model these components and aggregate them
to make predictions.


Hence, we use a temporal module with a Bahdanau attention mechanism in
charge of study seasonality and trend of the series, a spatial module in which
we propose a new spatio-temporal attention mechanism to model short-term and
spatial relations, and a dense module for retrieving and joining both previous
modules together with autoregressive terms and exogenous data in an unique
prediction. While we expect spatial and temporal modules to use inertia
information too, we reinforce this component with autoregressive terms as deep
neural models has shown lack of ability in modelling it (see Section
\ref{S3.2.3}).

Thanks to their capability to provide extra information about the network
intra-operation and feature importance, interpretability and explainability are
growing in importance and relevance. As we are specially interested in
demonstrating that CRANN modules have the behaviour just discussed,
interpretability is notably useful in our case. Concretely, attention mechanisms
are gaining supporters thanks to their capability of achieving good performance,
generalizing, and introducing a natural layer of interpretability to the
network. Thus, both temporal and spatial modules are covered. The dense module
makes use of SHAP values for estimating how important is each component for the
final prediction.

In order to showcase the proposed forecasting framework, the problem of traffic
intensity prediction is tackled in this paper. This real world problem
represents a perfect example of long, high-frequency time series which are
spatially interrelated, highly chaotic and with a clear presence of the four
aforementioned classical time series components. Furthermore, its environmental,
economic, and social importance turns it into a very relevant problem in need of
operational and cheap solutions.

In fact, with the increase of vehicles all over the world, several complications
have appeared recently: from traffic jams and their impact on economy and air
quality, going through traffic accidents, and health-related issues, to name a
few. Owing to the relevance of the matter, intelligent transport systems have
arised as an important field for the sake of improving traffic management
problems and establishing sustainable mobility as a real option. As an immediate
consequence, traffic prediction can be considered as a crucial problem on its
own and a perfect candidate as a real application that could benefit from
adaptable, accurate and interpretable NNs. For example, these kind of systems
might help to improve route-recommendation systems by not only estimating but
predicting, to optimize in real time buses waiting times, and to extract better
spatio-temporal information that would be helpful for traffic planning and
management. Although traffic systems are usually focused on short-term
forecasting\footnote{Traffic forecasting is commonly classified as short-term if
  the prediction horizon is less than 30 minutes and long-term when it is over
  30 min. We adopt that terminology throughout this paper.}, for academic
purposes we tackle the long-term problem by predicting 24 hours in order to
demonstrate that our model is capable of learning intrinsic spatio-temporal
traffic dependencies and patterns. However, as we will show, the model is easily
adaptable to any forecast window.



The main contributions of this study are summarized as follows:
\begin{itemize}
\item A new deep neural network especially designed for spatio-temporal
  prediction is proposed.
\item A novel spatio-temporal attention-based approach for regression is presented.
\item The contribution is illustrated by tackling a traffic prediction problem
  which is considered hard in both dimensions.
\item Results show that our proposal beats other state-of-the-art models in
  accuracy, adaptability and interpretability.
\end{itemize}

The rest of the paper is organized as follows: related work is discussed in
Section \ref{S2}, while Section \ref{S3} presents the problem formulation and
our deep learning model for spatio-temporal regression. Then, in Section
\ref{S4} we introduce our dataset, experimental design and its properties.
Section \ref{S5} illustrates the evaluation of the proposed architecture as
derived after appropriate experimentation. Finally, in Section \ref{S6} we point
out future research directions and conclusions.

\section{Related work}
\label{S2}

\subsection{Deep neural networks for spatio-temporal regression}
\label{S2.1}

Classic statistical approaches and most of the machine learning techniques that
are used to deal with spatio-temporal forecasting sometimes perform poorly due
to several reasons. Spatio-temporal data usually presents inherent interactions
between both spatial and temporal dimensions, which makes the problem more
complex and harder to deal with by these methodologies. Also, it is very common
to make the assumption that data samples are independently generated but this
assumption does not always hold because spatio-temporal data tends to be highly
self correlated.

On the contrary, models based on deep learning present two fundamental
properties that make them more suitable for spatio-temporal regression: their
ability to approximate arbitrarily complex functions and their facility for
feature representation learning, which allows for making less assumptions and
permits the discovery of deeper relations in data.

Within deep learning, almost all type of networks have been tried for
spatio-temporal regression. The most common ones are recurrent neural networks
(RNN), which due to its recursive structure have a privileged nature for working
with ordered sequences as time series. Nevertheless, it is not easy to use them
to model spatial relations, which makes them less suitable for this kind of
problems. For this reason, RNN models are usually combined with some spatial
information, as convolutions or spatial matrices. Previous works within the RNN
group are \cite{wang_predrnn_2017, alahi_social_2016, tang_st-lstm_2019} for
example. While RNN have received a lot of attention during last years, interest
in convolutional neural networks (CNN) for spatio-temporal series is recently
growing. Not only these systems are capable of exploiting spatial relations, but
they are showing state-of-the-art performance in extracting short-term temporal
relations too. For example, \cite{ke_hexagon-based_2019,
duan_deep_nodate,zhu_wind_2018} propose the use of CNN in spatio-temporal
regression.

In recent years, more complex models based on both RNN and CNN are replacing
traditional neural networks in this kind of problems. This is the case of
sequence to sequence models (seq2seq) and encoder-decoder architectures. By
enlarging the input information into a latent space and correctly decoding it,
these models have induced a boost in spatio-temporal series regression. As it
happened with RNN, spatial information is usually introduced explicitly. Some
examples might be found in \cite{liao_dest-resnet_2018,
  liao_deep_2018,yang_optimized_2017}. Finally, attention mechanisms were
introduced by \cite{luong_effective_2015, bahdanau_neural_2016} for natural
language processing. However, some researches have recently shown their ability
to handle all kind of sequenced problems, as time and spatio-temporal series.
Particularly, they have demonstrated to be a promising approach in capturing the
correlations between inputs and outputs while including a natural layer of
interpretability to neural models. This attention mechanisms might be introduced
at any dimension: spatial \cite{ChengSZH18}, temporal \cite{chen_novel_2020} or
both of them \cite{fu_spatiotemporal_2019}.



For a survey that recopilates the main characteristics of deep learning methods for
spatio-temporal regression and a vast compilation of previous work, see
\cite{wang_deep_2019}.


\subsection{Traffic prediction}
\label{S2.2}

Traffic flow prediction has been attempted for decades, and has experienced a
strong recent change after the emerging methodologies that let us model
different traffic characteristics. With the increase of real-time traffic data
collection methods, data-based approaches that use historical data to capture
spatio-temporal traffic patterns are every day more common. We will divide this
data-driven methods into three major categories: statistical models, general
machine learning models and deep learning models. 

Within statistical methods, the most successful approach has been ARIMA and its
derivates, which have been used for short-term traffic flow prediction
\cite{hamed_mohammad_m_short-term_1995}. Afterwards, some expansions as KARIMA
\cite{voort_combining_1996} and SARIMA \cite{kumar_short-term_2015} were also
proposed to improve traffic prediction performance. Nevertheless, these models
are constrained according to several assumptions that, in real world data as
our, do not always fit properly.

Amongst general machine learning approaches, bayesian methods have shown to
adapt well when dealing with spatio-temporal problems \cite{sun_trac_nodate,
  queen_intervention_2009, pascale_2011_nodate}, as their graph structure fits
in a road-network visualization. However, they do not always show better
performance when compared to other methodologies that will be presented next.
Another usual technique in the field of forecasting spatio-temporal series are
tree models. Within this area, there are different approaches
\cite{dong_short-term_2018, alajali_intersection_2018, alajali_traffic_2018_2},
each one with its own advantages and disadvantages. Generally tree models are
easily interpretable, making them a good option if the main interest is to
better understanding the phenomena. Nevertheless, tree models tend to overfit
when the amount of data and dimensions of the problem is big, as it normally is
the case in traffic prediction. As with trees, support vector machines (SVM) and
support vector regression (SVR) have been widely used, as in
\cite{wu_multiple_2008, cong_traffic_2016, mingheng_accurate_2013}. While SVM
and SVR perform well, these methodologies must establish a kernel as a basis for
constructing the model. This means that, for such an specific problem like the
one we are working on, the use of a predetermined kernel (usually radial) might
not be flexible enough.

In a closer line with our work, deep learning has been widely used for traffic
forecasting. The idea of stacking CNNs modules over LSTMs (or vice versa) is
usual in recent literature. Some of the most interesting work in this category
applied to traffic prediction can be found in \cite{liu_urban_2018,
  ermagun_spatiotemporal_2018}. Furthermore, \cite{wu_hybrid_2018} shows that
the combination of this modules together with an attention mechanism for both
space and time dimension, might be beneficial. In this same line,
\cite{do_effective_2019} proposes a spatio-temporal attention mechanism and show
how through interpretability we can extract valuable information for traffic
management systems. Lately, other options have been considered as using 3
dimensional CNNs for making the predictions \cite{guo_deep_2019} to effectively
extract features from both spatial and temporal dimensions or combining CNN-LSTM
modules with data reduction techniques in order to boost performance
\cite{bogaerts_graph_2020}. In \cite{deng_exploring_2019} authors present an
example of how to deal with incomplete data while still being capable of
exploring spatio-temporal traffic relations. As it was mentioned before, the
vast majority of these works have a set of fixed conditions and mainly focus on
short-term predictions. Longer-term predictions (with horizons of more than
fours hours) can also be found in \cite{qu_daily_2019} in which a neural
predictor is used to mine the potential relationship between traffic flow data
and a combination of key contextual factors for daily forecasting, and
\cite{he_stcnn_2019} where ConvLSTM units try to capture the general
spatio-temporal traffic dependencies and the periodic traffic pattern in order
to forecast one week ahead.


In concordance with these last works, our model is designed to be adaptable to
both long and short term forecasting. Also, it is not limited nor evaluated
over a set of fixed conditions, letting us extract more general conclusions.

\subsection{Time series decomposition in deep neural models}
\label{S2.3}

Time series decomposition and derived methods for regression have been widely
studied in the statistical context. Beyond standard methodologies, as ARIMA and
exponential smoothing, more elaborated proposals have been suggested. For
example, in \cite{bergmeir_bagging_2016} a bootstrap of the remainder for
bagging several time series via exponential smoothing is proposed. Similarly,
\cite{de_oliveira_forecasting_2018} presents an extension of an analogous
methodology using SARIMA. In both cases, its demonstrated that a proper use of
time series components for modelling can be profitable and thus this remains as
a promising research line.

In the concrete case of deep neural models, although using time series
decomposition in order to improve and boost the performance of deep neural
networks is not new, most of previous research has focused on using those
components externally to the network. Several studies point out that, before
feeding the network, it might be beneficial to detrend the series to just build
a prediction model for the residual series \cite{li2014comparison,
  dai_deeptrend_2017, dai_deeptrend_2019}. Other works show how autoregressive
methods together with deep neural models help to tackle the scale insensitive
problem of artificial neural networks \cite{lai_modeling_2018} and allow for the
implementation of several temporal window sizes for training efficiently
\cite{liu_traffic_nodate}. Deseasonalisation in order to to minimise the
complexity of the original time series has been recommended through several
works too \cite{bandara_lstm-msnet_2019, BENTAIEB20127067, nelson_time_1999}.


However, the way in which neural networks relate to time series components
remains an open issue. Although it has been demonstrated that a correct
decomposition of the series can help the system, it is not clear how deep neural
models can deal with these components by themselves without the need of external
information as we propose.

\subsection{Interpretability}
\label{S2.4}

Defined as the ability to explain or to present in understandable terms aspects
of a machine learning algorithm operation to humans, interpretability is growing
in importance, specially in deep neural models due to its black-box nature.
Until now, it has principally been investigated and demonstrated in a wide
variety of tasks such as natural language processing, classification
explanation, image captioning, etc \cite{bahdanau_neural_2016,li_deep_nodate,
  You_2016_CVPR}. However, in regression problems and particularly in traffic is
still an open issue and there is a long way to go. For example,
\cite{WANG2019372} demonstrates that by using a bidirectional LSTM that models
paths in the road network and analyzing feature from the hidden layer outputs it
is possible to extract important information about the road network. Similarly,
\cite{cui_high-order_2019} studied the importance of the different road segment
when forecasting traffic via a graph convolutional-LSTM network. In general,
traffic interpretability research has focused in pointing out important road
segments. On the contrary, \cite{do_effective_2019} presents a comprehensive
example of spatio-temporal attention in which both dimensions are analysed from
an interpretability point of view, and not only from the spatial one.

Following this last idea, we propose a methodology in which spatio-temporal
interpretability is taken into account and, at the same time, go deeper in
understanding how important these two dimensions are to model the generator
process of our problem.

\section{CRANN Model and problem formulation}
\label{S3}


\subsection{Problem formulation}
\label{S3.1}

Given a spatial zone $S$, where each traffic sensor is represented as $s_i$, and
a timestep $t_j$, we aim to learn a model to predict the volume of traffic in
each sensor $s_i$ during each time slot $t_j$. This mean that a spatio-temporal
sample writes as ${x_{s_i,t_j} : j = 1, . . . , T; i = 1, . . . , S}$. From now
on, we will distinguish a prediction from a real sample by using
$\tilde{x}_{s_i,t_j}$ for the first one.

\subsection{CRANN: a combination approach for spatio-temporal regression}
\label{S3.2}


As stated above, CRANN model is based in the idea of combining neural modules
with the intention to exploit the various components that can be identified in a
spatio-temporal series: seasonality, trend, inertia and spatial relations. By
combining different neural architectures focused on each component, we expect to
avoid redundant information flowing through the network and to maximize the
benefits of each approach. As a result, several layers of interpretability will
allow us to better understand the problem being modelled and to verify that our
model is working the way we were expecting.

The code for the software used in this paper can be found in \textit{\url{https://github.com/rdemedrano/crann_traffic}}.

\subsubsection{Temporal module}
\label{S3.2.1}

There is a consensus that dealing with long-term sequences using ordinary
encoder-decoder architectures is a promising approach. However, the fact that
only the final state of the encoder is available to the decoder limits these
models when trying to make short or long-term predictions
\cite{cho-etal-2014-properties}.
Particularly, in traffic we would expect an improvement in performance when
taking into account not only closer states of the desired output, but also 
several days.

In order to solve this problem, several encoder-decoder architectures that use
information from some or all timesteps have been proposed. Among all these
models, a particular approach has shown good qualities by improving performance
and adding an interpretability layer to the system: attention mechanisms.
Presented in several ways \cite{luong_effective_2015, bahdanau_neural_2016}, the
idea behind these mechanisms lies in creating a unique mapping between each time
step of the decoder output to all the encoder hidden states. This means that for
each output of the decoder, it can access to the entire input sequence and can
selectively pick out specific elements from that sequence to produce the output.
In other words, for each output, the network learns to pay attention at those
past timesteps (inputs) that might have had a greater impact in the prediction.
Typically, these mechanisms are exemplified by thinking of manual translation:
instead of translating word by word, context matters and it is better to focus
more on specific past words or phrases to translate the next.

Following that rationale, our temporal module is formed by two LSTMs working as
encoder and decoder respectively. The first one inputs the time series and
outputs a hidden state $s$, while the second one inputs a concatenation of
attention mechanism output $c$ (named 'context vector') and the previous decoder
outputs, and uses this information to perform its prediction.

As it can be seen, the structure is very similar to a sequence to sequence model
without bottleneck but with the introduction of new information via the
attention mechanism. The idea behind this model is explained below. For
simplicity, notation is coherent with the one used by Bahdanau
\cite{bahdanau_neural_2016}
through this section. For each forecast step $i$, the context vector
is calculated taking into account the encoder hidden state for each input
timestep $j$:
\begin{equation}
    c_{i} = \sum_{j = 1}^{N} \alpha_{i,j} h_{j}, \label{eq:1}
\end{equation}
where $N$ is the input sequence dimension (coincident with number of encoder
hidden sates), $\alpha_{i,j}$ is the attention weight defined as how much from
the encoder hidden state $j$ should be payed attention to when making the
prediction at time $i$. It is computed as follows:
\begin{equation}
    \alpha_{i,j} = \frac{f(h_{i}, s_{j})}{\sum_{j = 1}^{N} f(h_{i}, s_{j})}. \label{eq:2}
\end{equation}
In this last expression, $h$ is the decoder hidden state and $f$ refers to an
attention function that estimates attention scores between $s$ and $h$.
Depending on the attention mechanism, many functions have been suggested as
attention functions (for example, dot products, concatenation, general...). In
this work, a feedforward neural network that combines information from both the
encoder and the decoder is chosen. Specifically, it writes:
\begin{equation}
    f(h_{i}, s_{j})  = W_{c} \cdot tanh(W_{d} \cdot h_{i} + W_{e} \cdot s_{j}), \label{eq:3}
\end{equation}
where $W$s are weight matrices.

Finally, the new decoder hidden state $h'_{i}$ is obtained through concatenating
$c_{i}$ with $h_{i}$ and the output can be decoded as
\begin{equation}
    h'_{i}  = [c_{i}; h_{i}]. \label{eq:4}
\end{equation}

\begin{figure}[tbp]
\centering
\includegraphics[width=0.5\textwidth]{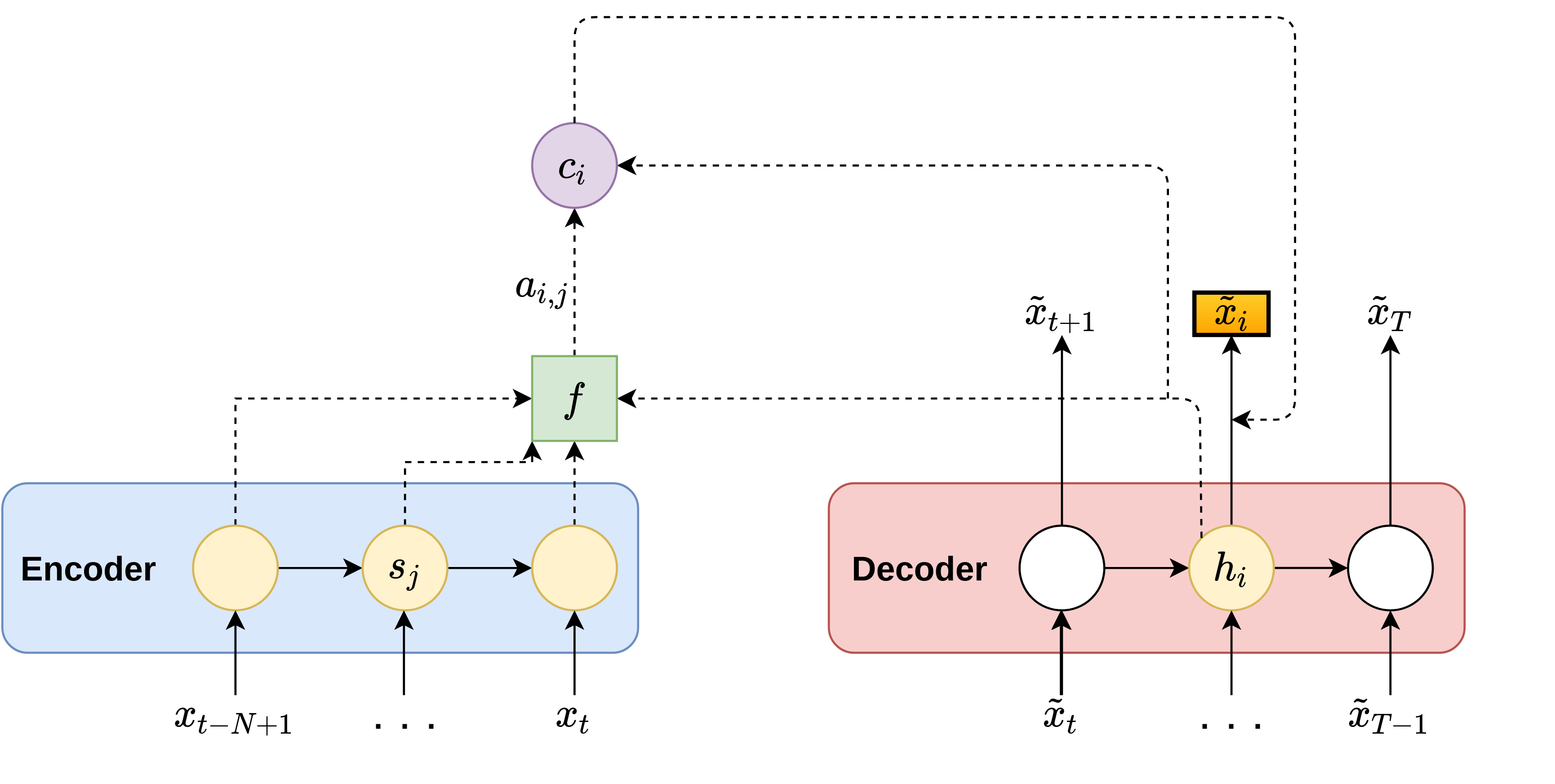}
\caption{Schematics of the attention mechanism used in our temporal module when
  predicting timestep $i$ for a $T$ horizon forecasting. In yellow, hidden
  states for both encoder and decoder that are used to compute the prediction.
  In green, the attention function. In violet, the obtained context vector based
  on attention weights $\alpha_{i,j}$ and $h_{i}$. In orange, the predicted
  value computed by concatenating $c_{i}$ with $h_{i}$.}
\label{fig:att_mech}
\end{figure}

The complete process is summarised in Fig. \ref{fig:att_mech}. The temporal
module of CRANN focuses its effort on discovering and modelling long-term time
relations of the complete system by using average traffic for the complete zone.
From equations (\ref{eq:1}-\ref{eq:4}), it should be clear that no spatial
relations have been explored or introduced. Although it might seem more
profitable to capture these relations for each spatial series, we would be
learning redundant knowledge once that the spatial module comes out. By taking
into account information from several past weeks, the model will be capable of
capturing the periodicity and trend changes in the series, which might be
fundamental for a more precise forecasting. As traffic trend is not exactly
equal for a temporal window of several hours or days, it is necessary to adapt
CRANN temporal module input to an amount of time that let us avoid temporal
information loss. In particular, we will use two weeks as input for a 24 hour
output.

\subsubsection{Spatial module}
\label{S3.2.2}

Even though traffic seems highly dependent on its temporal dimension, it is also
clear that spatial relations are relevant. The premise of spatio-temporal
forecasting is based on not only taking into account that these relations exist,
but effectively using them to improve performance. In this context,
convolutional neural networks (CNN) appear as a perfect choice as they are meant
to precisely exploit spatial characteristics and interactions. Furthermore, as
it was mentioned in Section \ref{S2}, CNN are also gaining attention as a
promising paradigm to study short-term temporal associations. Hence, we propose
a novel spatio-temporal attention mechanism that tackles two major aspects of
spatio-temporal forecasting with CNN: adds a new layer in order to improve
spatial relations and our understanding over them in a specific problem, and
lets the CNN explore further temporal information.

As in the temporal case, this mechanism can be introduced as a new layer through
the network, meaning that our system will consist in an usual CNN followed by
the spatio-temporal attention mechanism. The CNN will enrich input information
and compute some output $x_{\mathrm{conv}}$ with same dimensions as its input.
In other words, it will be the one in charge to improve the quality of the input
while making sure to keep some aspects of the original structure of the series.
In some way, it is equivalent to the encoder model from Section \ref{S3.2.1},
except that 
there is not an equivalent structure as the own attention mechanism can handle
it. This mechanism works by assigning a score 
$\sigma_{i,j,k}$ to every pair of spatial points $(j,k)$ for each
input lag $i$. The score $\sigma_{i,j,k}$ represents how important is the point
$k$ in lag of the input $i$ in order to calculate the prediction for point $j$ for
all output timesteps. It writes as:
\begin{equation}
  \sigma = g(x_{\mathrm{conv}}, W_{\mathrm{att}}), \: W_{\mathrm{att}} \in \mathbb{R}^{T \times S \times S}, \label{eq:5}
\end{equation}
where $T$ is the number of timesteps, $S$ the number of spatial points, $g$ is
an attention function that calculates an attention score and $W_{\mathrm{att}}$
defines the spatio-temporal attention tensor. $W_{\mathrm{att}}$ is a learnable
tensor which can be interpreted as a means to modelate spatio-temporal
interdependencies of the system. It can be decomposed in a three-dimensional
space, meaning that $W_{\mathrm{att}}^{i,j,k}$ encodes how does the point $j$ at
timestep $i$ interact with the CNN output $x_{\mathrm{conv}}$ to make the
prediction.

Given that each element of $W_{\mathrm{att}}$ is expected to provide information
about the system dynamics, the attention function $g$ is useful to modulate
concrete relations, element by element, for a given input series. Thus, it is
defined as follows:
\begin{equation}
    g(x_{\mathrm{conv}}, W_{\mathrm{att}}) = x_{\mathrm{conv}} \circ W_{\mathrm{att}}, \label{eq:6}
\end{equation}
where $\circ$ is the Hadamard product (also known as the element-wise, entrywise or Schur 
product)
Although some other functions as concatenation and a feedforward neural network have been
tested, no improvement was reported. Moreover, Hadamard product stands out for
its simplicity and for offering a naive explanation about the inner functioning
of the spatio-temporal attention mechanism.

Once that these attention scores have been calculated, it is preferable to give
them some properties that ease the interpretation and convergence of the
attention mechanism. The spatio-temporal attention matrix $a \in \mathbb{R}^{T
  \times S \times S}$ is defined as a three-dimensional tensor that meets the
following conditions:
\begin{itemize}
\item Each attention weight is constrained between zero and one: $a_{i,j,k} \in
  [0,1]$.
\item As $a_{i,j,k}$ represents the importance of point $k$ at timestep $i$ to
  predict $j$, the sum of attention weights for each timestep $i$ and point $j$
  must add up to one: $\sum_{k = 1}^{S} a_{i,j,k} = 1$.
\end{itemize}

By enforcing these conditions we can therefore infer a probabilistic
interpretation of the attention weights. This can be done by applying a softmax
operator over the third dimension:
\begin{equation}
    a = \mathrm{softmax}(\sigma). \label{eq:7}
\end{equation}

\begin{figure*}[hbtp]
  \centering
  \includegraphics[width=1\textwidth]{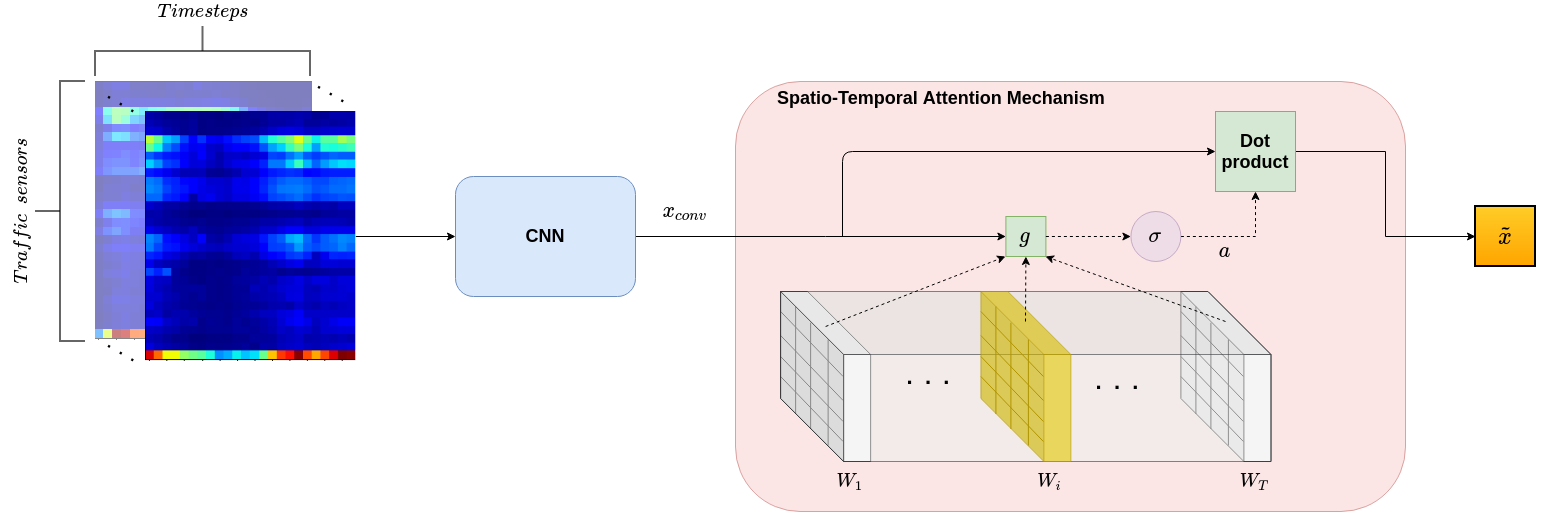}
  \caption{Schematics of the attention mechanism used in our spatial module when
    predicting all timesteps for a $T$ horizon forecasting given a batch of
    inputs. In yellow, a representation of $W_{\mathrm{att}}$ for an input
    timestep $i$. In green, the attention function and dot product. In violet,
    the obtained score vector $\sigma$ based on $x_{\mathrm{conv}}$ and
    $W_{\mathrm{att}}$. In orange, the predicted value.}
  \label{fig:spatial_att}
\end{figure*}

Finally, in order to calculate the definitive prediction $\tilde{x}$, we use the
inner product between tensors. This way we can easily interpret the output as a
weighted sum over all spatio-temporal input conditions through the attention
weights. Depending on how relevant each input element is for the regression, it
will contribute in a different manner to the final output:
\begin{equation}
    \tilde{x} = a \cdot x_{\mathrm{conv}}.  \label{eq:8}
\end{equation}

The complete process is summarised in Fig. \ref{fig:spatial_att}. To conclude,
the spatial module of CRANN focuses its efforts on discovering and modelling
spatial and short-term time relations of the complete system by using real
traffic data for each sensor. For the CNN, a 2D model in which every channel
corresponds to a timestep is used. The architecture consists of a sequence of
convolutions layers, batch normalization and ReLU activation. In particular, we
will use a 24 hour input for a 24 hour output.

\subsubsection{Dense module and training}
\label{S3.2.3}

At this point, on the one hand we have modeled the general behaviour of all the
involved time series and we thus have trend information from the temporal
module, and on the other hand we have explored spatial relations and specific
predictions for each traffic sensor through the spatial model. Hence, it is
necessary to join both modules in some way that let us exploit all the available
information for the sake of improving the final performance. At the same time,
it might be interesting to introduce available exogenous knowledge that might
affect the future of the series. While several exogenous variables are well
known as important for traffic forecasting, we will only use meteorological
features as we reckon they might be enough to prove how our model works when
using exogenous data in a first approximation. Lastly, since inertia has a
central role in time series forecasting, we include the timesteps $t-1$ to $t-4$
as autoregressive terms. Although we might expect that both previous modules
take into account this inertia at some level, as Ling et al.
\cite{lai_modeling_2018} pointed out, \textit{due to the non-linear nature of
  the convolutional and recurrent components, one major drawback of the neural
  network model is that the scale of outputs is not sensitive to the scale of
  inputs}, meaning that in real datasets with severe scale changing like ours
this effect might be problematic. Thus, making use of this information directly
is also expected to benefit the final performance.

\begin{figure*}[htbp]
\centering
\includegraphics[width=0.75\textwidth]{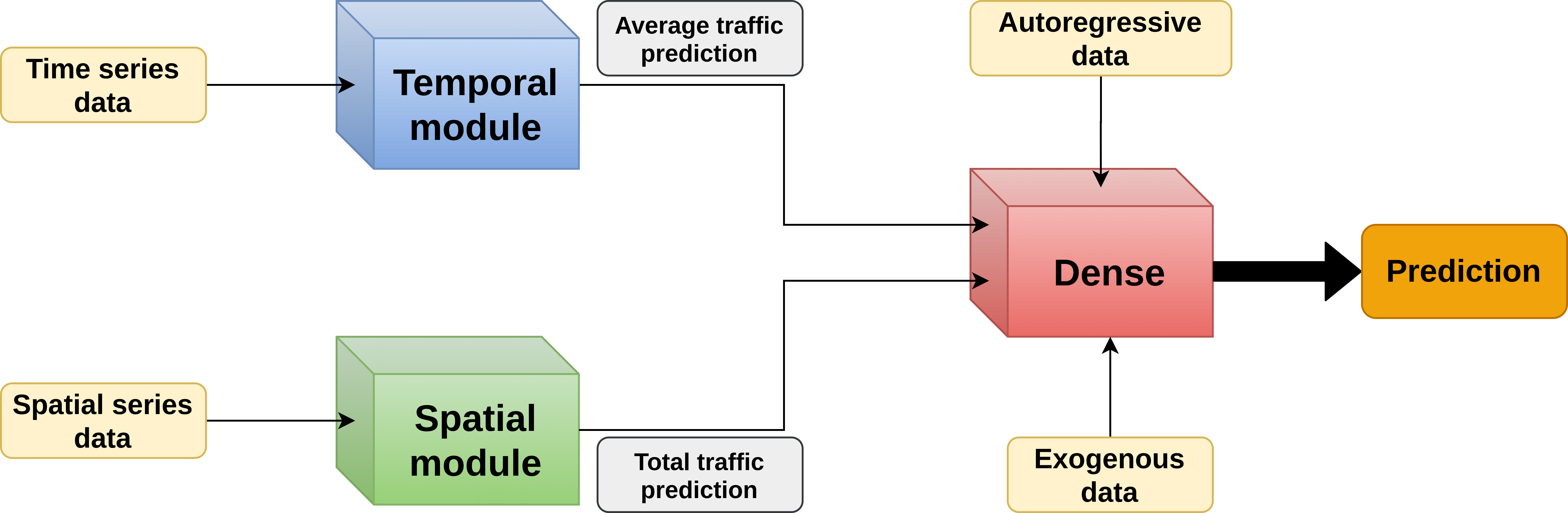}
\caption{Schematics for the CRANN architecture. In yellow, input data sources.
  Time series data refers to average traffic historic data, spatial series data
  to the real traffic historic, and exogenous data are in this case weather
  predictions. In grey, output from the different middle stages. In orange, the
  final spatio-temporal prediction.}
\label{fig:model_total}
\end{figure*}

The resulting CRANN architecture is shown in Fig. \ref{fig:model_total}. It
consists of a dense module whose inputs are the exogenous data, the
autoregressive data and the outputs of both the temporal and spatial previously
described modules. This final dense module is simply a fully connected
feedforward neural network.

While all these modules could be stacked, we decided to use a mixed
parallel/series structure for the sake of improving modularity and
explainability through the network. By having a compendium of models with a
specific and clear job working independently, it is easier to train, improve,
remodel or change any of them if needed. Moreover, the stacked approach was
tested but no significant accuracy improvement was reported.

Regarding to training, it can be done at once or in several steps (for each
module) in order to parallelize the process. Also, all weights are randomly
instantiated using ``Xavier'' initialization \cite{glorot_understanding_nodate}.
Finally, and independently on how the network is being trained, the training can
be summarised as with any other neural network as searching some parameters
$\theta^{*}$ via minimization of a cost function $L$:
\begin{equation}
\theta^{*} =  \mathop {{{\,\mathrm{arg\,min}\,}}}\limits _{\theta } L(\theta) \label{eq:9}
\end{equation} 

\subsubsection{Interpretability}
\label{S3.2.4}

The ability to interpret the trained models is nowadays a must-have in every
machine learning research check-list. For that reason, we should value
methodologies that are able to offer explanations about their predictions. In
the particular case of CRANN, interpretability has been put into practice as
follows:
\begin{itemize}
\item Temporal module: By using a temporal attention mechanism, we have an
  intrinsic interpretability layer. Since we defined attention weights as how
  important each lag from the input sequence is for predicting each output
  timestep (see Section \ref{S3.2.1} for a deeper insight), we can easily
  interpret these weights to better understand how is our temporal module making
  use of the inputs when forecasting.
\item Spatial module: As with the temporal module, the underlying attention
  mechanism provide and easy and natural interpretation. Attention weights
  typify how significant is every spatial point when predicting in the
  spatio-temporal domain. Furthermore, they might be represented by input
  lag or aggregated.
\item Dense module: As the information flowing through the previous modules has
  a clear interpretation (the temporal module outputs the average traffic for
  the whole space and the spatial module outputs actual spatio-temporal
  predictions), it is straightforward to interpret the network with several
  feature analysis methods (like integrated gradients
  \cite{sundararajan2017axiomatic}) or saliency methods (like SHAP values
  \cite{lundberg2017unified}). 
  In this work, SHAP values are chosen. It is
  important to remark that with a non-parallel join of modules (Section
  \ref{S3.2.4}), this methodologies might not be as convenient due to
  non-explainable inputs of the dense module, i.e., by having interpretable
  middle stages through the network is easier to elucidate if certain
  information is contributing to the final prediction or not.
\end{itemize}


\section{Data and experiments}
\label{S4}

To characterize and validate our proposed model, this section provides
information related to all the decisions taken and the experiments performed. As
explained above, we focus our work on the long-term forecasting problem, i.e.,
a 24 hours spatio-temporal prediction.

\subsection{Data description and analysis}
\label{S4.1}

In order to validate the CRANN framework for spatio-temporal forecasting, we
chose the problem of predicting traffic intensity in the city of Madrid. The
data available came from two different sources:
\begin{itemize}
\item \textbf{Traffic data:} Provided by the Municipality of Madrid through its
  open data portal\footnote{\textit{Portal de datos abiertos del Ayuntamiento de
      Madrid}: \url{https://datos.madrid.es/portal/site/egob/}\label{note1}},
  this dataset contains historical data of traffic measurements in the city of
  Madrid. The measurements are taken every 15 minutes at each point, including
  traffic intensity in number of cars per hour
  . 
  Spatial information is given by traffic sensors with their coordinates
  (longitude and latitude). While a dense and populated network of over 4.000
  sensors is available, we decided to simplify and use only a selection of them,
  as explained below.

\item \textbf{Weather data:} Weather data was also provided by the Municipality
  of Madrid\footref{note1}. Weather observations consist of hourly temperature
  in Celsius degrees, solar radiation in W/m\textsuperscript{2}, wind speed
  measured in ms\textsuperscript{-1}, wind direction in degrees, daily rainfall
  in mmh\textsuperscript{-1}, pressure in mbar, degree of humidity in percentage
  and ultraviolet radiation in mWm\textsuperscript{-2} records. Weather
  information is reported hourly and they are used as if they were numerical
  weather predictions (feeding the model at each moment with the data
  corresponding to the forecasting horizon).
\end{itemize}
In this work, only data from 2018 and 2019 is used.

\begin{figure}[tbp]
  \centering
  \begin{subfigure}[t]{0.5\columnwidth}
    \includegraphics[width = 1\textwidth]{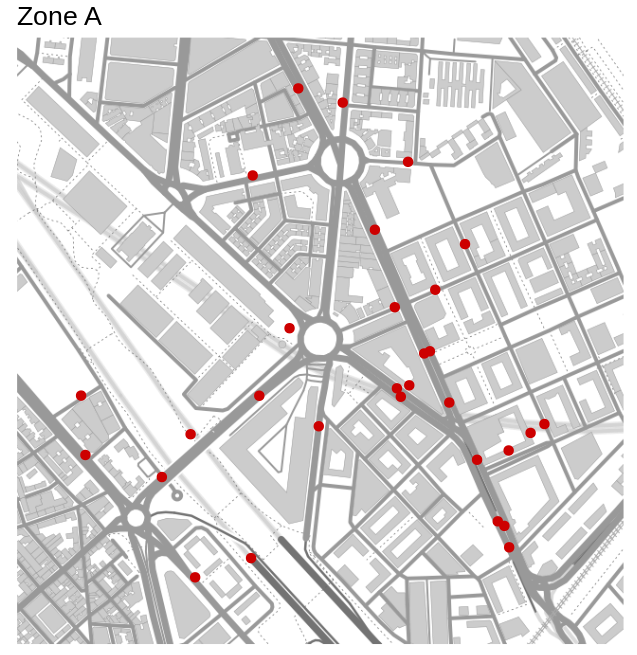}
    \label{fig:zone_a}
  \end{subfigure}%
    ~
  \begin{subfigure}[t]{0.5\columnwidth}
    \includegraphics[width = 1\textwidth]{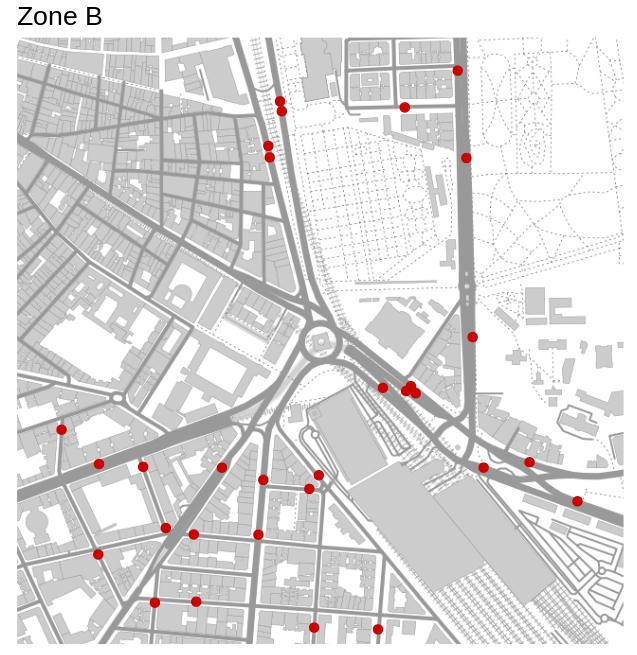}
    \label{fig:zone_b}
  \end{subfigure}
    \\
  \begin{subfigure}{0.5\columnwidth}
    \centering
    \includegraphics[width = 1\textwidth]{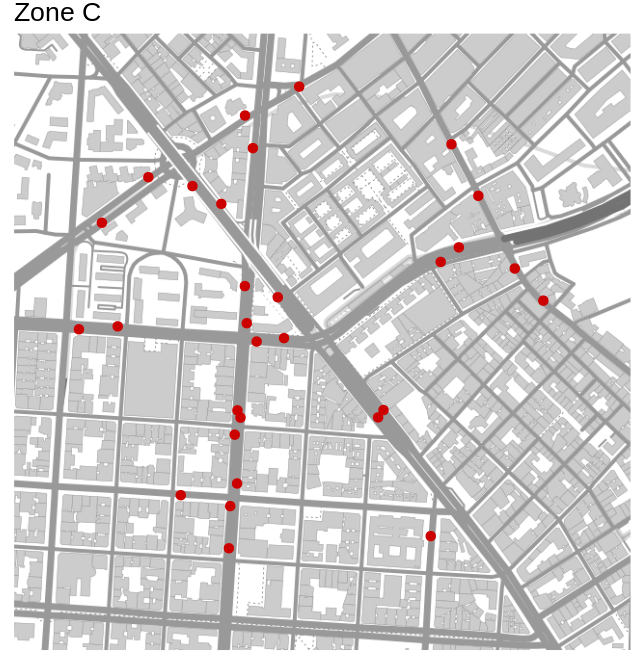}
    \label{fig:zone_c}
   \end{subfigure}%
~
  \begin{subfigure}{0.5\columnwidth}
    \centering
    \includegraphics[width = 1\textwidth]{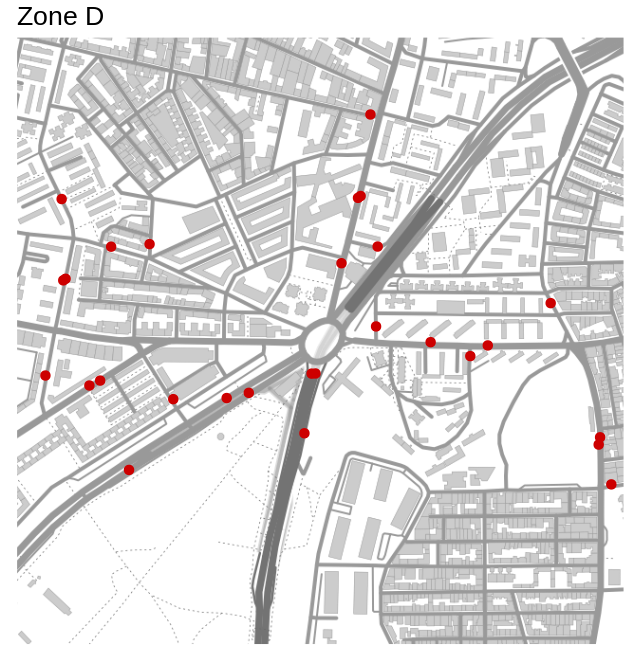}
    \label{fig:zone_d}
  \end{subfigure}
  
  \caption{Location of traffic sensors for each zone.}
  \label{fig:spatial_zones_sensors}
\end{figure}

For a more robust evaluation of the different models, four specific zones are
chosen (see table \ref{tab:spatial_zones}),
each one of them containing 30 traffic sensors (Fig.
\ref{fig:spatial_zones_sensors}). All these four zones are characteristic for
being hot spots of traffic in Madrid. In addition, they all present a wide
variety of traffic conditions: one-way streets, avenues, highways, roundabouts
and, in general, ways with different flow conditions. Statistics presented in table 
\ref{tab:spatial_zones} for each zone point in this direction. Although these spatial
dispositions result in a more complicated environment, makes our work more
general.


\begin{table}[btp]
\centering
\begin{tabular}{p{0.4cm}p{2.5cm}p{1cm}p{1cm}p{0.5cm}p{0.5cm}}
  \toprule
  Zone & Name & Longitude & Latitude  & Mean & Std \\
  \midrule
  A & Legazpi & -3.6952 & 40.3911 & 563.6 & 803.2 \\
  B & Atocha & -3.6920 & 40.4087 & 680.9 & 769.7 \\
  C & Avenida de América & -3.6774 & 40.4374 & 459.3 & 476.6 \\
  D & Plaza Elíptica & -3.7176 & 40.3852 & 360.2 & 517.8 \\ \bottomrule
\end{tabular}
\caption{Location of the center of spatial zones and name correspondence from
  now on. Main data statistics.
  }
\label{tab:spatial_zones}
\end{table}

Missing values are scarce (about 1$\%$ per series). They are replaced by sensor,
hour and day of the week aggregation as interpolation and closeness replacement
leads to greater loss of information. Outliers represents less than $0.001\%$ of
each series and are given by public events (for example, Champions League final
or Basketball World Cup). As these kind of events are not representative of our
problem, and thus they are excluded from our analysis.

The data are aggregated into 1-hour intervals and, due to the lack of outliers,
normalized using a min-max technique to the range [0,1]. Normalization constants
are calculated over the training dataset. Each spatio-temporal series is
normalized separately as we are looking for an agnostic scale for each sensor.

\begin{figure}[tbp]
  \centering
  \begin{subfigure}[t]{0.5\columnwidth}
    \includegraphics[width=4.2cm]{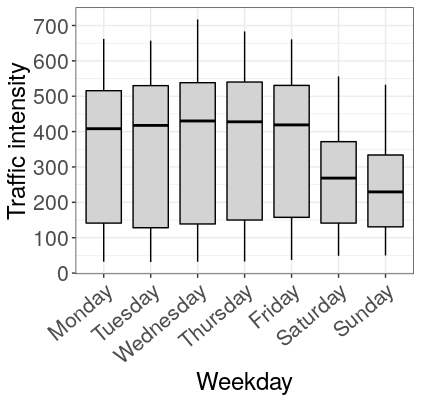}
    \label{fig:day_dist}
  \end{subfigure}%
  ~
  \begin{subfigure}[t]{0.5\columnwidth}
    \includegraphics[width=4.2cm]{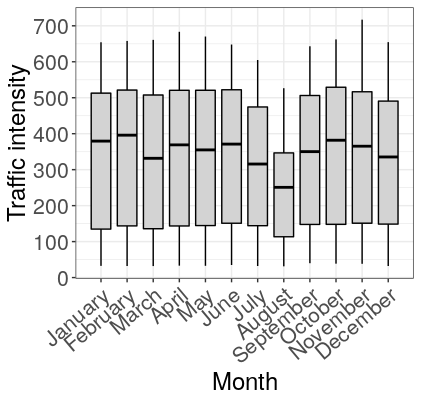}
    \label{fig:month_dist}
  \end{subfigure}
  \\
  \begin{subfigure}{\columnwidth}
    \centering
    \includegraphics[width=1\textwidth]{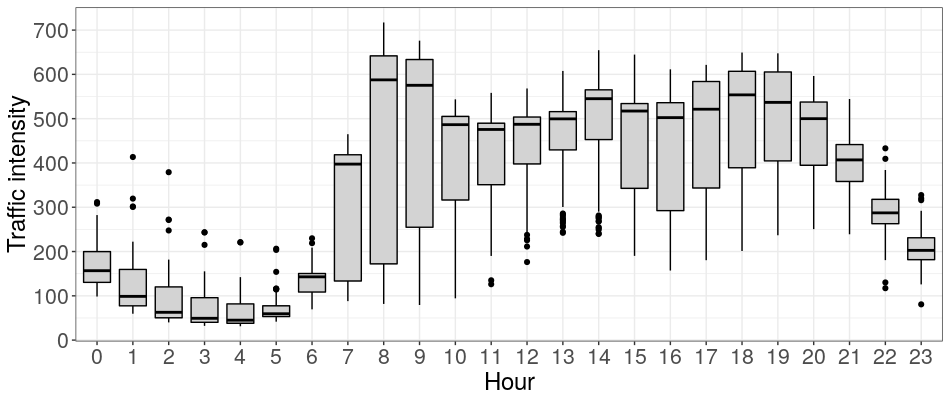}
    \label{fig:hour_dist}
  \end{subfigure}
  \caption{\label{fig:temp_dist} Hourly, weekly and monthly distribution of
    traffic intensity series.
    }
\end{figure}
\begin{figure*}[tbp]
\centering
\includegraphics[width = 1\textwidth]{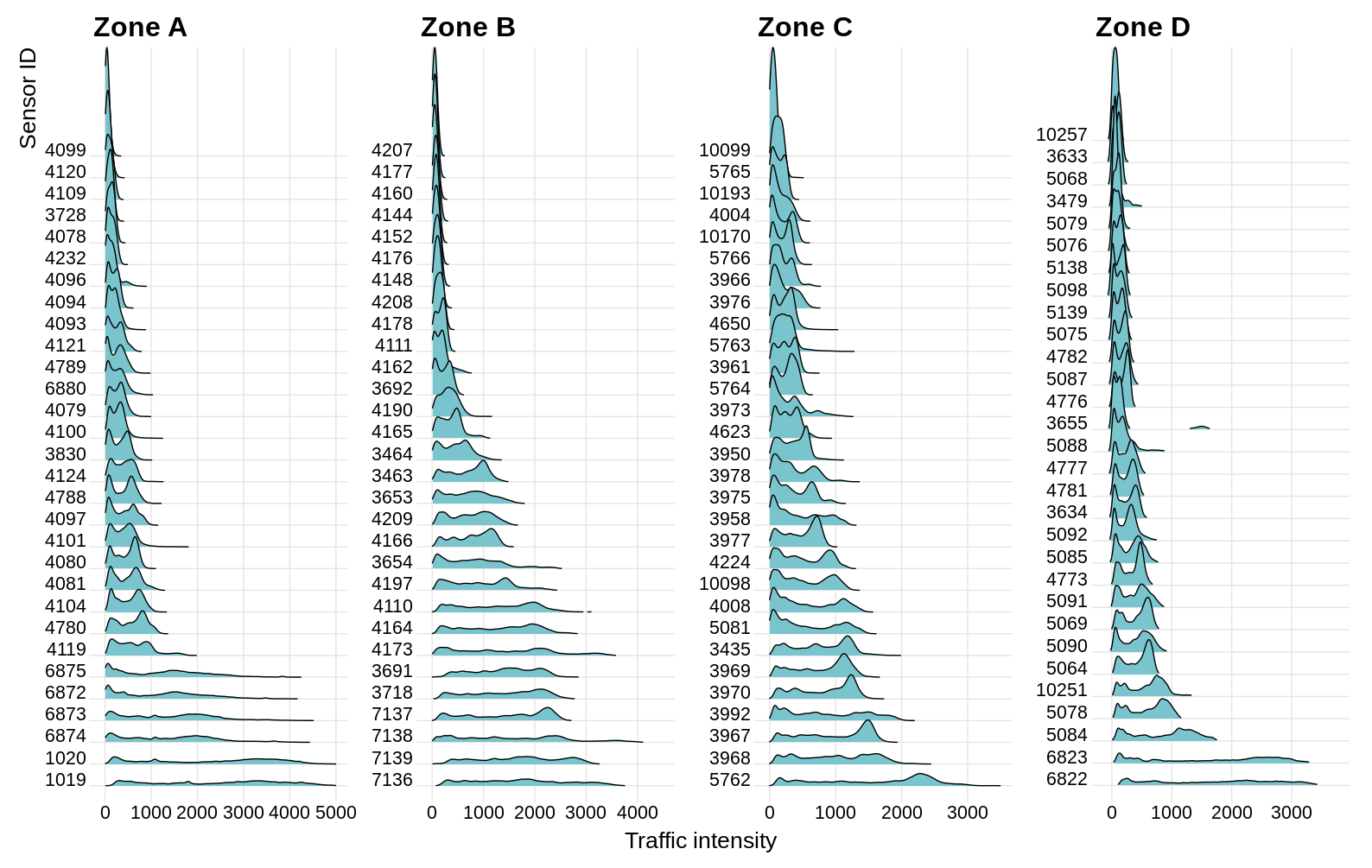}
\caption{Traffic intensity distribution by sensor (in number of vehicles per hour).}
\label{fig:spatial_dist}
\end{figure*}

In order to better understand our problem, we show significant properties of the
data. Due to high number of sensors and the spatial heterogeneity commented above, instead 
of showing general attributes from our series (as mean, median or dispersion) it is more
instructive to see both spatial and temporal distributions. Thus, in one hand,
Fig. \ref{fig:temp_dist} shows a boxplot for different time variables. From this
figure, it should be clear that traffic is highly dependent of time and periods
of human activity. On the other hand, the spatial distribution of our series is
displayed in Fig. \ref{fig:spatial_dist}. This last figure not only let us
better understand our data, but also reinforces the idea of having very diverse
spatial zones for our study. 

\subsection{Benchmark models}
\label{S4.2}


We compare the performance of the proposed CRANN with a CNN, an LSTM, the usual
combination CNN+LSTM and a sequence to sequence model (seq2seq).
\begin{itemize}
\item CNN: A 2D convolutional model in which every channel corresponds to a
  timestep. The architecture consists of a sequence of convolutional layers,
  batch normalization and ReLU activation. For every layer a kernel size of 3
  for each dimension is used. It uses 24 hours as input and outputs a 24 hours
  prediction.

\item LSTM: These models have several hidden layers with a number of hidden
  units to determine. We used the $\tanh$ activation functions as in the
  original model. The number of inputs and outputs are equivalent to the number
  of sensors. Although GRUs modules have also been tested, no difference has
  been reported. It uses historical data from two weeks as input and outputs a
  24-hour prediction.
    
\item CNN+LSTM: A stacked model consisting of a CNN module whose output is in
  turn the input of a LSTM. Both modules are defined as the two previous models.
  It uses 24 hours as input and outputs a 24-hour prediction.

\item seq2seq: These architectures are based on an encoder and a decoder, both
  LSTMs, without ``bottleneck''. That is to say, hidden variables from all
  timesteps are used as inputs for the decoder. The number of inputs and outputs
  are equivalent to the number of sensors. As with LSTMs, GRUs did not show a
  better performance. It uses two weeks as input and outputs a 24 hours
  prediction.
\end{itemize}

With these models and their implementation particularities (inputs and outputs)
we aim to cover a wide range of neural network paradigms for our comparison. For
instance, CNN are specially designed for learning spatial relations, LSTMs and
Seq2Seq models are designed to explore mainly time interactions and CNN+LSTM
are closer to our model being a mixture of both previous approaches.


Concerning hyperparametrization and training, instead of using preset
architectures, to be fair, the optimal configuration for each model was obtained
via Bayesian hyperparameter optimization \cite{wu_hyperparameter_2019} which is
defined as: building a probability model of the objective function and using it
to select the most promising hyperparameters to evaluate in the true objective
function. Unlike grid search and random search, the bayesian approach keeps
track of past evaluation results. Final hyperparameters can be found in table
\ref{tab:params}.

\begin{table}[btp]
  \centering
  \begin{tabular}{llcc}
    \toprule
    {model} & {hyperparameter} & {value} & {\# parameters} \\ \toprule
    \multirow{1}{*}{CNN}    & Convolutions  & (32,32,32,64,64,64) & \multirow{1}{*}{132k} \\ \midrule
    
    \multirow{2}{*}{LSTM}   & Number of layers & 2  &  \multirow{2}{*}{206k} \\ 
                            & Hidden units    & 100   \\ \midrule
    
    \multirow{3}{*}{CNN+LSTM} & Convolutions &  (32,32,64,64,64) & \multirow{3}{*}{329k} \\ 
                            &  Number of layers &  2 \\ 
                            & Hidden units & 100 \\  \midrule
    
    \multirow{2}{*}{seq2seq} & Number of layers & 2 & \multirow{2}{*}{368k} \\
                            & Hidden units &  100 \\  \midrule
    
    \multirow{4}{*}{CRANN} & Convolutions &  (64,64,64,64,64) & \multirow{4}{*}{1M} \\ 
                            & Number of layers &  1 \\ 
                            & Hidden units & 100 \\ 
                            & Dense layers & 1 \\ \bottomrule
  \end{tabular}
  \caption{Values used for each hyperparameter and total number of parameters.
    }
  \label{tab:params}
\end{table}

All the models are trained using the mean squared error (MSE) as objective
function with the Adam optimizer \cite{kingma_adam_2017}. The batch size is 64,
the initial learning rate is 0.01 and both early stopping and learning rate
decay are implemented in order to avoid overfitting and improve performance. The experiments run in a NVIDIA RTX 2070.

\subsection{Experimental design}
\label{S4.3}


In order to guarantee that models can be compared in a fair manner it is
essential to fix the approach to error estimation, which must be shared as much
as possible by all models.
First of all, as stated in \cite{bergmeir_note_2018}, standard
$k$-cross-validation is the way to go when validating neural networks for time
series if several conditions are met. Specifically, that we are modelling a
stationary nonlinear process, that we can ensure that the leave-one-out
estimation is a consistent estimator for our predictions and that we have
serially uncorrelated errors.

While the first and the third conditions are trivially fulfilled for our
problem, the second one needs to be specifically studied for the sake of
avoiding data leakage. Given that we use all the possible series, even though
the ones are unrepeated, it is possible to introduce prior information from the
training to the test via closeness of samples (for example training a sequence
whose start is at 10:00 AM and testing in a sequence whose start is at 11:00 AM
from the same day i.e. one timestep forward). Due to this problem, it is not
possible to create random folds and it is necessary to specify a separation
border among different sets (training, validation and test).

\begin{figure*}[tbp]
  \centering
  \includegraphics[width = 0.75\textwidth]{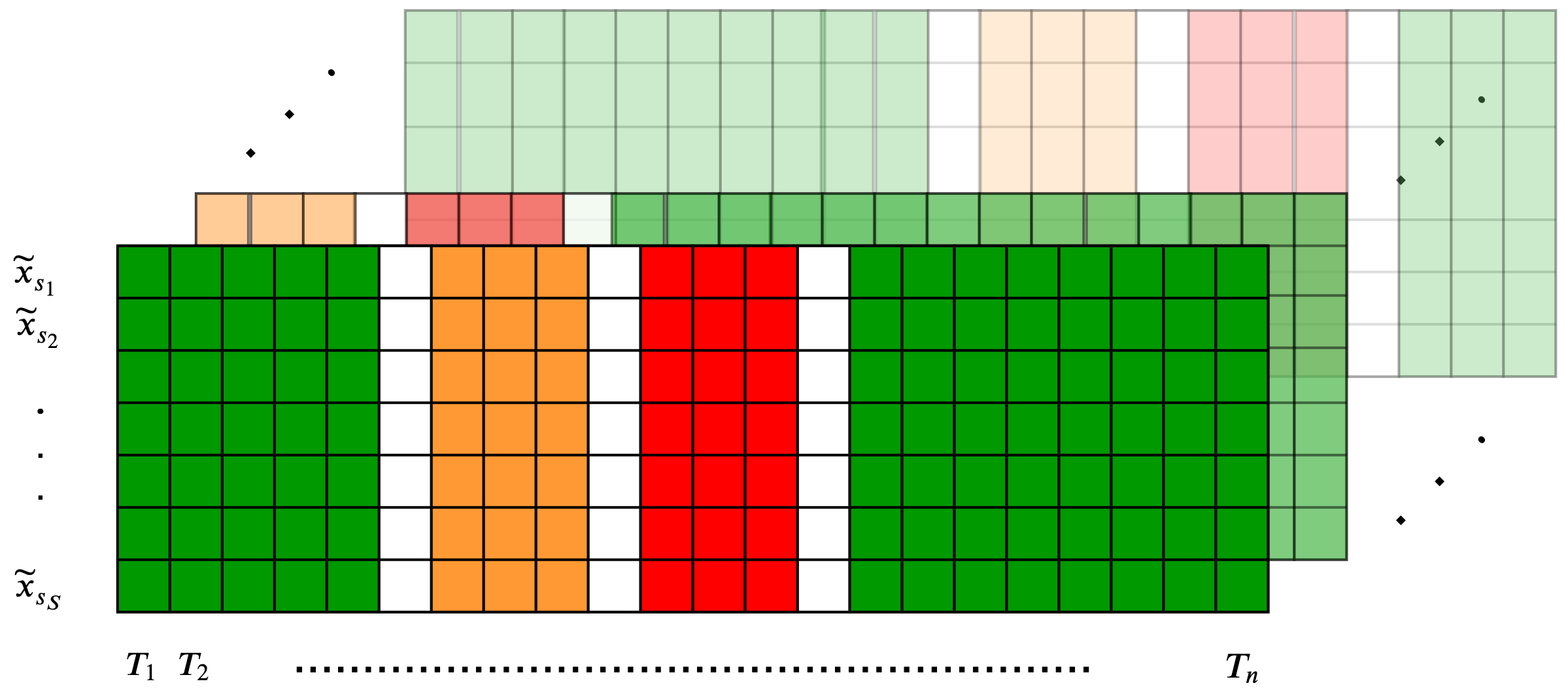}
  \caption{Validation methodology example with training (green), validation
    (orange) and test (red) sets for our proposed cross-validation procedure.
    Rows shown in white are omitted due to dependency considerations.}
  \label{fig:val}
\end{figure*}

In this concrete case, this separation takes as much timesteps as every model
uses for its training. A scheme of this methodology is shown in Fig.
\ref{fig:val}. Particularly, a 10-cross-validation scheme without repetition is
used for each spatial zone separately.

To evaluate the precision of each model, we computed root mean squared error
(RMSE), bias and weighted mean absolute percentage error (WMAPE). In a
spatio-temporal context \cite{wikle_spatio-temporal_2019}, they are defined as:
\begin{equation}
  \mathrm{RMSE} = \sqrt{\frac{1}{TS} \sum_{j = 1}^{T} \sum_{i = 1}^{S} (\tilde{x}_{s_i,t_j} - x_{s_i,t_j})^{2}}, \label{eq:10}
\end{equation}

\begin{equation}
  \mathrm{bias} = \frac{1}{TS} \sum_{j = 1}^{T} \sum_{i = 1}^{S} ( \tilde{x}_{s_i,t_j} - x_{s_i,t_j} ), \label{eq:11}
\end{equation}

\begin{equation}
  \mathrm{WMAPE} = 100 \times \frac{\sum_{j = 1}^{T} \sum_{i = 1}^{S} \mid \tilde{x}_{s_i,t_j} - x_{s_i,t_j} \mid}{\sum_{j = 1}^{T} \sum_{i = 1}^{S} \mid x_{s_i,t_j} \mid}, \label{eq:12}
\end{equation}
where (as in Section \ref{S3.1}) ${x_{s_i,t_j} : j = 1, . . . , T; i = 1, . . .
  , S}$ is a spatio-temporal sample from the real series, $\tilde{x}_{s_i,t_j}$
represents the predicted series, $S$ is the total number of traffic sensors and
$T$ the total number of predicted timesteps.

For all these metrics, the closer to zero they are the better the performance
is.

\section{Results}
\label{S5}

\subsection{Error estimation}
\label{S5.1}

A general comparison for the different error metrics can be seen in Table
\ref{tab:res_total}. Bias is represented by its absolute value. These values
correspond to averaging each metric for all spatial zones. Highlighted in bold,
CRANN results shows a better performance overall for all errors.

\begin{table}[btp]
  \centering
  \begin{tabular}{lS[table-format=3.2]S[table-format=2.2]S[table-format=2.2]S[table-format=2.2]}
    \toprule
    {model}   &  {RMSE}  & {$|\mathrm{bias}|$} & {WMAPE} & {Run time (s)} \\ \midrule
    CNN & 238.24 & 22.12 & 25.89 & 68 \\ 
    LSTM & 255.76 & 19.58 & 27.46 & 552 \\ 
    CNN+LSTM & 252.34 & 21.70 & 27.29 & 144 \\ 
    Seq2Seq & 246.45 & 19.14 & 25.79 & 1098 \\ 
    CRANN &  $\mathbf{221.31}$  & $\mathbf{17.80}$ & $\mathbf{23.18}$ & 1083 \\ \bottomrule
  \end{tabular}
  \caption{Average performance for $t = 1$ to $t = 24$, calculated over all
    spatial zones and average run time per fold. For a more detailed view of error metrics distribution, see
    Fig. \ref{fig:res1}.}
  \label{tab:res_total}
\end{table}

\begin{figure*}[tbp]
  \centering
  \includegraphics[width = 1\textwidth]{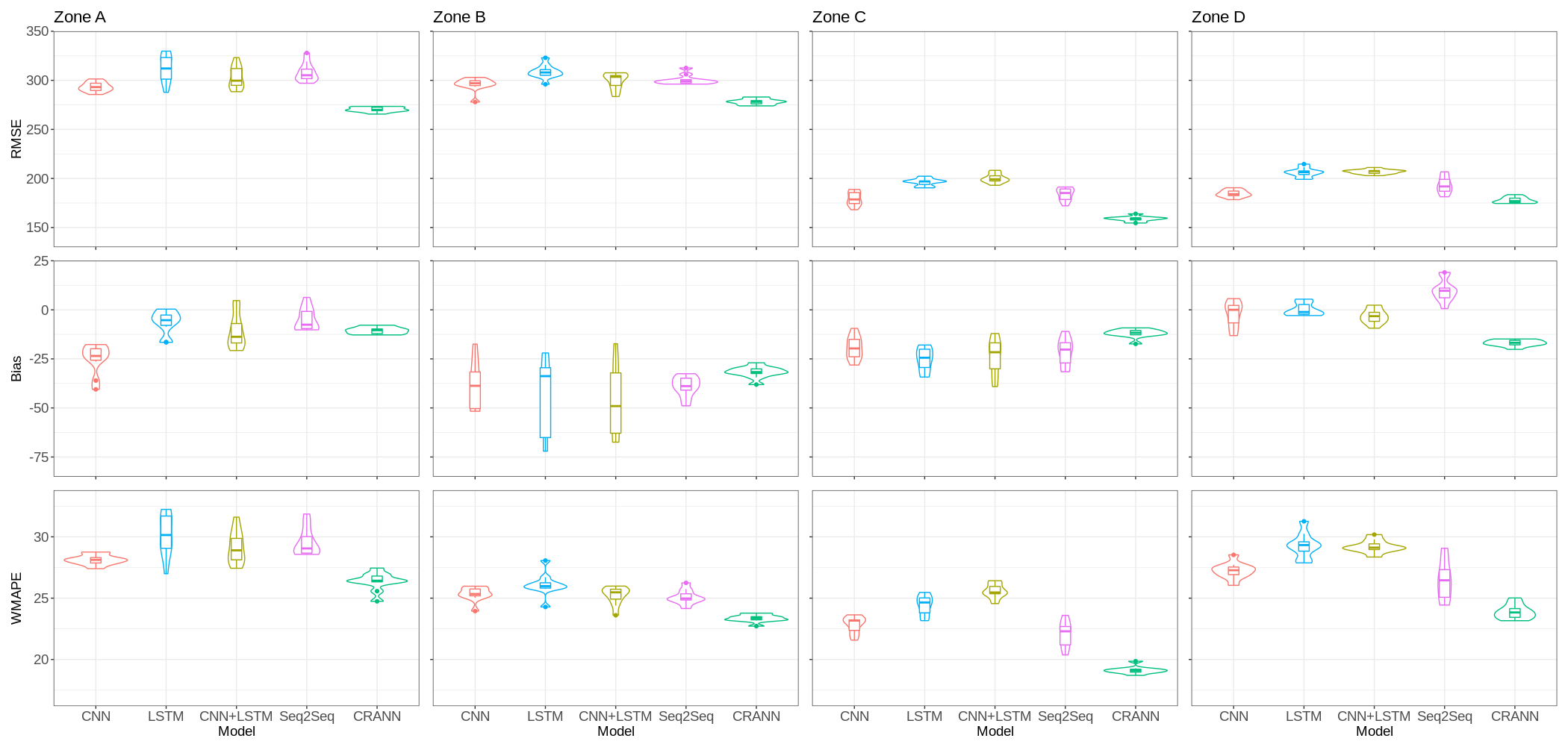}
  \caption{Violin plots of error metrics distributions for each zone and model.}
  \label{fig:res1}
\end{figure*}

For a better understanding on how each model is performing, Fig. \ref{fig:res1}
present these same error metrics but with their distribution for each zone
separately. While LSTM and recurrent models in general are a standard for time
series forecasting, our experiments demonstrate that standard CNN can perform
similar (or even better) than recurrent models and should have a bigger space in
time series. Also, vanilla LSTM might not be the best option for a real world
spatio-temporal system with high complexity. Oddly, the CNN+LSTM model performs
worse than the traditional CNN model, which can be due to the LSTM module
negatively affecting its behaviour. 
With p-values $<$ 0.05 when comparing with all the baseline models, CRANN can be considered as
statistically significantly better at all error metrics with a confidence of
$95\%$.

From Fig \ref{fig:res1} we can also deduce that the deviation of the CRANN model
is generally stable and is the smallest one. In fact, models that are highly
dependent on a recurrent neural network show a higher-deviation tendency respect
to strongly CNN-based models.

\begin{figure*}[tbp]
\centering
\includegraphics[width = 1\textwidth]{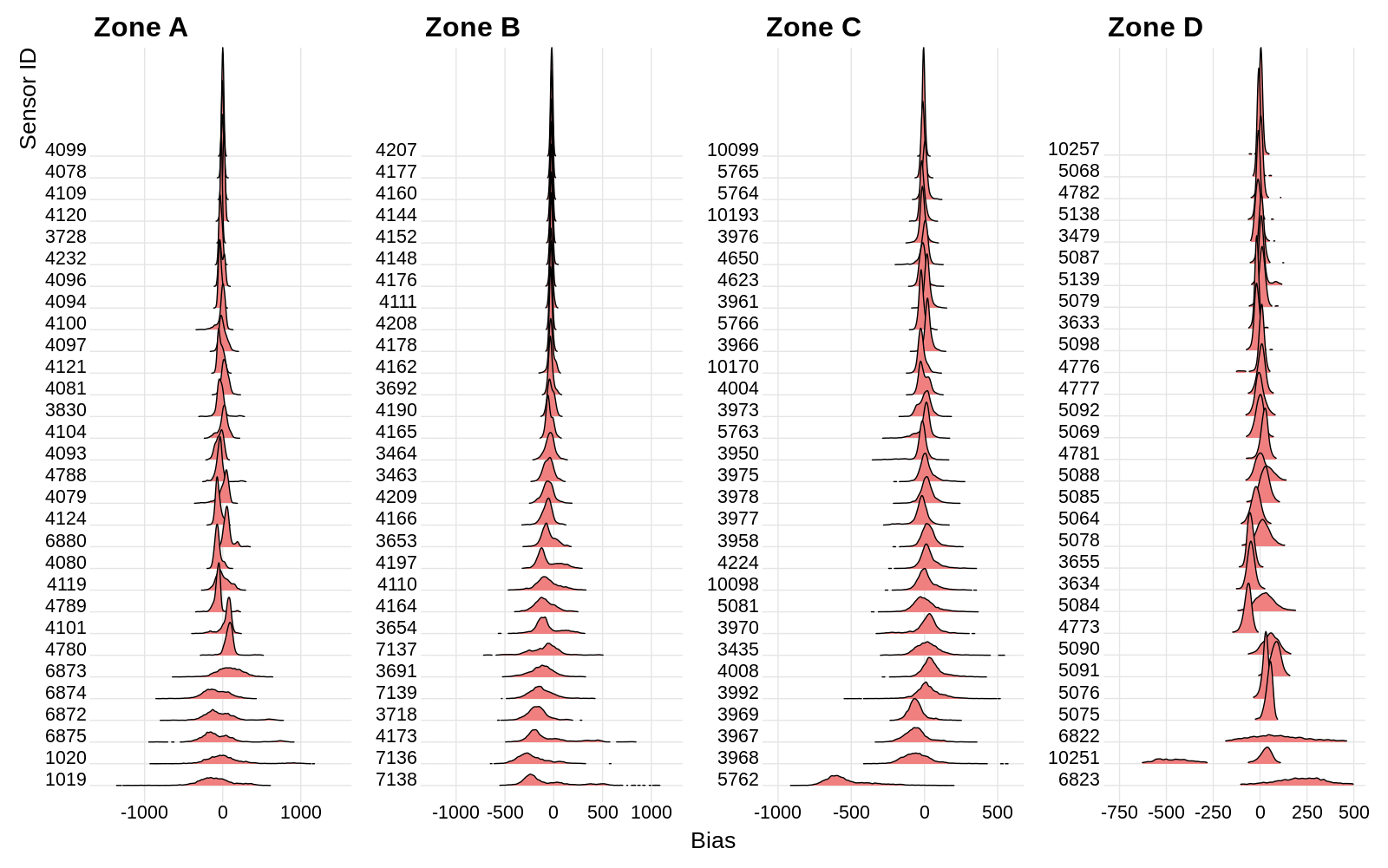}
\caption{Bias distribution for each traffic sensor. When compared with Fig.
  \ref{fig:spatial_dist} it should be clear that measurement points with higher
  traffic intensities and more variability are shifted to the left in bias,
  resulting in a underestimation of the real series.}
\label{fig:res3}
\end{figure*}

Bias exhibits a clear under zero tendency, meaning that all models tend to
underestimate their predictions. For a deeper understanding of this phenomenon,
Fig \ref{fig:res3} shows CRANN's bias spatial distribution for each studied
sensor. Compared with Fig \ref{fig:spatial_dist}, it is clear that traffic
sensors with higher traffic intensity values, which in turn coincide with
those sensors with distributions with greater dispersion, are mainly responsible
of this behaviour. While we would expect higher errors in these kind of sensors
with such an aggressive traffic pattern, it is not clear why the shifting occurs
in only one direction. Nevertheless, as this anomaly happens for all CRANN and
baseline models in every zone, we expect that its nature is intrinsic for the
system or the validation methodology.


\begin{figure}[tbp]
\centering
\includegraphics[width=1\columnwidth]{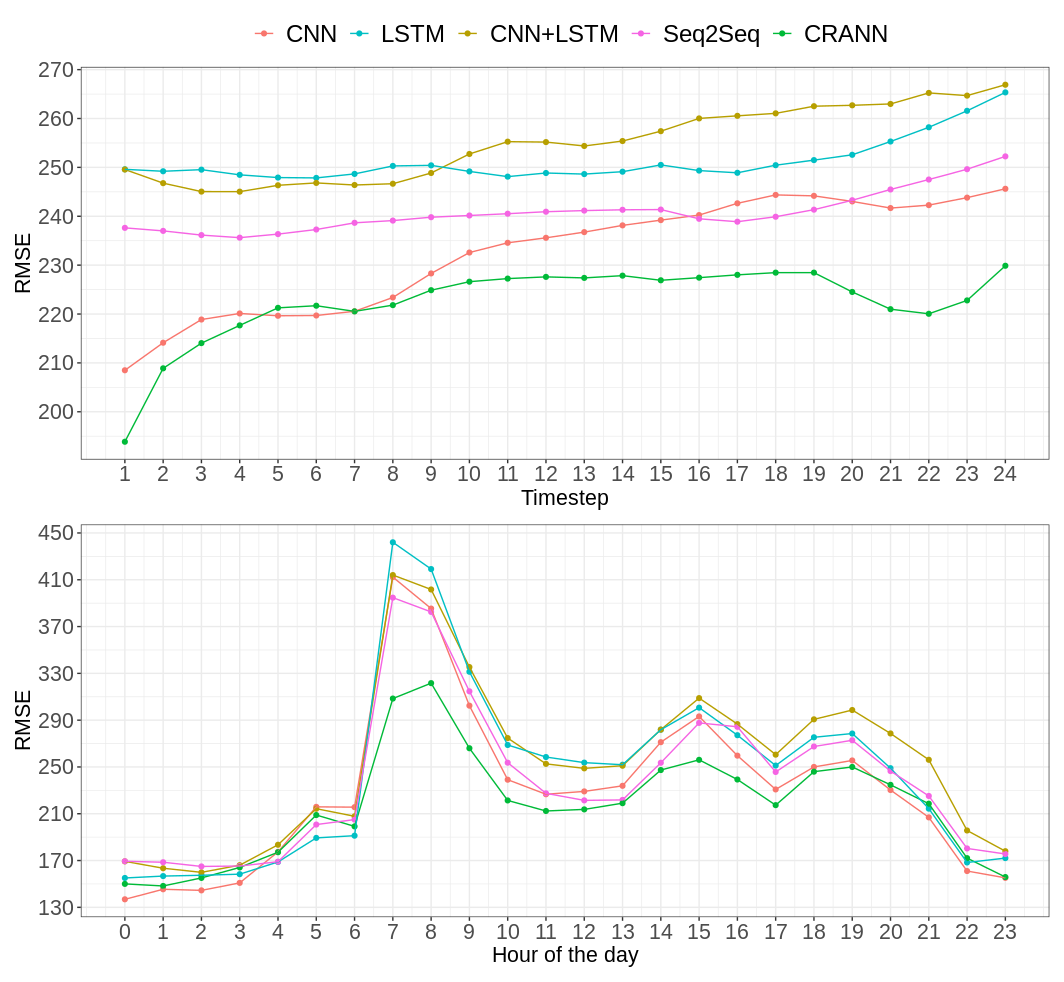}
\caption{RMSE analysis for time dimension. How it varies depending on the prediction timestep for all models (top). Average error depending on the our of the day (bottom).}
\label{fig:res4}
\end{figure}

With respect to time dimension, a simple analysis can show some expected
behaviour. As shown in Fig. \ref{fig:res4} (top), all models experiment an increase of
average RMSE when the predicted timestep goes further, as we could expect. As
spot-forecasting is based on evaluation through all possible series, these
timesteps do not have a direct correspondence with specific hours of the day and
this figure is not contaminated by natural dynamics of traffic.
However, there are two clear patterns: LSTM-based models (LSTM, CNN+LSTM and
seq2seq) share a higher error for the first horizons, which are usually
considered easier to predict under the hipothesis of inertia of the series. This
tells us that they are not capturing this inertia correctly. At the same time,
CNN-based models (CNN and CRANN) manage to capture the inertia of the series.
Having introduced autoregressive terms into the CRANN model stands as a positive
alternative to alleviate and improve this difficulty. Also, we can see a valley
from timestep $t+20$ to $t+24$ as due to traffic periodicity, that fraction of
the series is highly similar to the one introduced as autoregressive terms
(timesteps $t-1$ to $t-4$).

Meanwhile, Fig. \ref{fig:res4} (bottom) let us understand how the average error of the 
different models are distributed as a function of the hour of the day in which the
prediction is being made. As we would expect, these errors are bigger at rush
hours, giving us a distribution with same shape than the one presented in Fig.
\ref{fig:temp_dist}. Nevertheless, CRANN model stands out for its ability to
outperform significantly its rivals in those exact instants,
when it is precisely more useful and challenging to get a good behavior.

\begin{figure*}[tbp]
\centering
\includegraphics[width=1\textwidth]{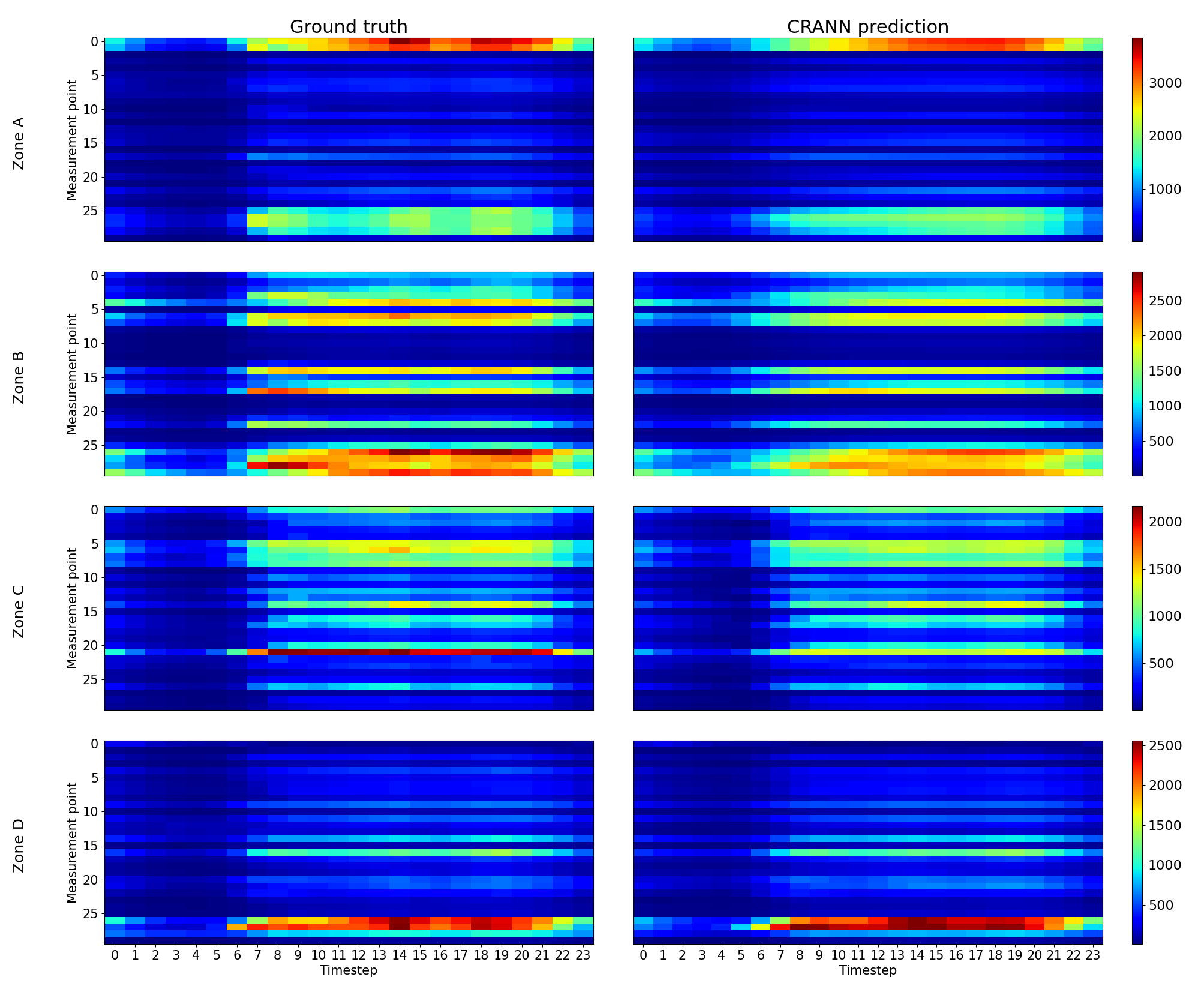}
\caption{Example of CRANN's predictions for all zones. Average results for
  predictions starting at 00:00 and ending at 23:00.}
\label{fig:res5}
\end{figure*}

Lastly, Fig. \ref{fig:res5} displays an example forecast of CRANN. By taking all
series starting at 00:00 and ending at 23:00 it is possible to visualize
average performance of the model in a specific context. This figure clearly shows
how our model is successfull in learning the spatio-temporal dynamic of
traffic, even adapting its behaviour to fine details in a highly complex
spatio-temporal problem.

\subsection{Interpretability}
\label{S5.2}

In order to better understand how our model works, we might use all the
interpretability layers presented in Section \ref{S3.2.4}. Also,
interpretability will let us corroborate our initial hypothesis about how each
module tackles different aspect of spatio-temporal series: trend, seasonality,
inertia and spatial relations.

\begin{figure*}[tbp]
\centering
\includegraphics[width=1\textwidth]{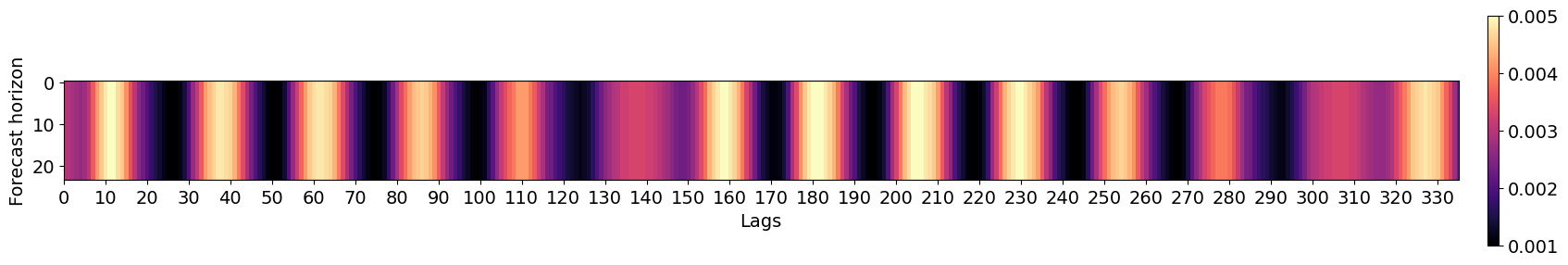}
\caption{Average temporal attention given by the temporal module of CRANN.
  Attention weights are represented as a function of input and output series. In
  the x-axis, past lags from the input series. In the y-axis, forecast horizons
  (i.e. future lags) from the output series. A $(x,y)$ value represents how
  important is timestep $x$ to predict timestep $y$.
    }
\label{fig:inter1}
\end{figure*}

Starting by our temporal module (see Sec. \ref{S3.2.2}), Fig. \ref{fig:inter1}
shows average attention weights computed by the attention mechanism in function
of both input and output timesteps. From this figure we can have a clear
intuition about the 24 hour pattern that our model has learned. At the same
time, time-back steps $~160$ and $~325$, which correspond to 7 and 14 days
before the prediction, show to be more important as traffic presents a seven
days seasonality too. As the input series approach to the forecasting window,
the importance keeps growing proving that the temporal module is regulating
trend as we were looking for. The fact that no shifting is happening is due to
averaging over all test samples.

\begin{figure}[tbp]
\centering
\includegraphics[width=1\columnwidth]{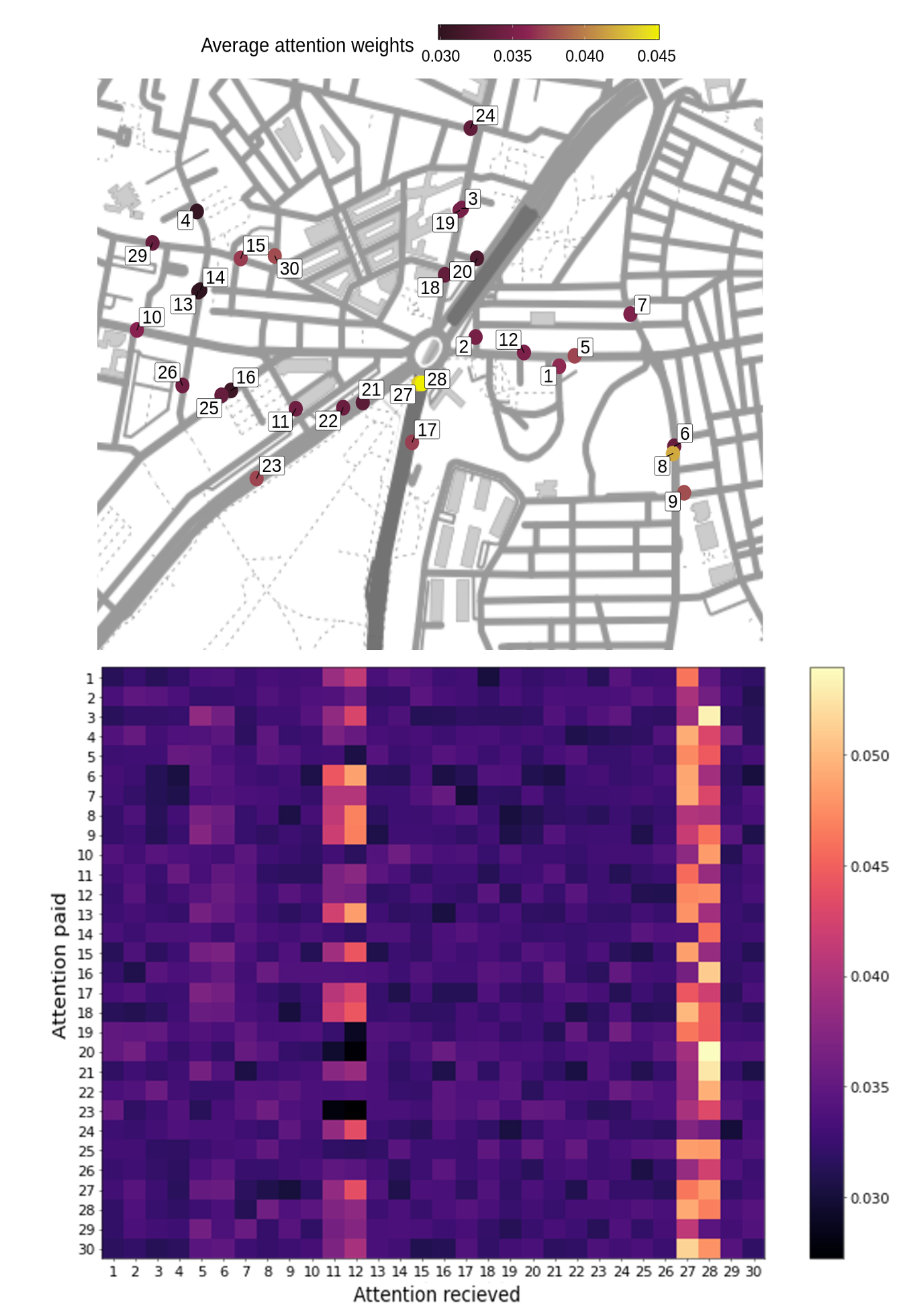}
\caption{Average spatial attention given by CRANN spatial module at zone D. Attention
  weights are averaged for all sensors and timesteps (top). Attention weights for each pair of sensors (bottom).}
\label{fig:inter5}
\end{figure}

With respect to the spatial module attention (see Sect. \ref{S3.2.3}), it is
obviously highly dependent on each specific zone. For that reason, Fig.
\ref{fig:inter5} illustrates the attention weights for traffic sensors in Zone
D. As our defined spatial attention mechanism uses different weights for each
lag of the input, average values are shown. As it can be seen (top), sensors with
specially complex conditions (high traffic intensity, big avenues...) are
usually scored as more important by the spatio-temporal attention mechanism.
This is the case of points 8, 27 and 28 for example. On the contrary, those
points that we would expect to have less impact in global traffic show smaller
values, like sensors 4, 16 and 20. Similarly (bottom), sensors in heavy traffic intensity emplacements show to receive higher attention. As we tackle the long-term forecasting problem, we do not expect our model to pay attention by closeness, but by general importance in the entire zone.

\begin{figure}[tbp]
  \centering
  \includegraphics[width=1\columnwidth]{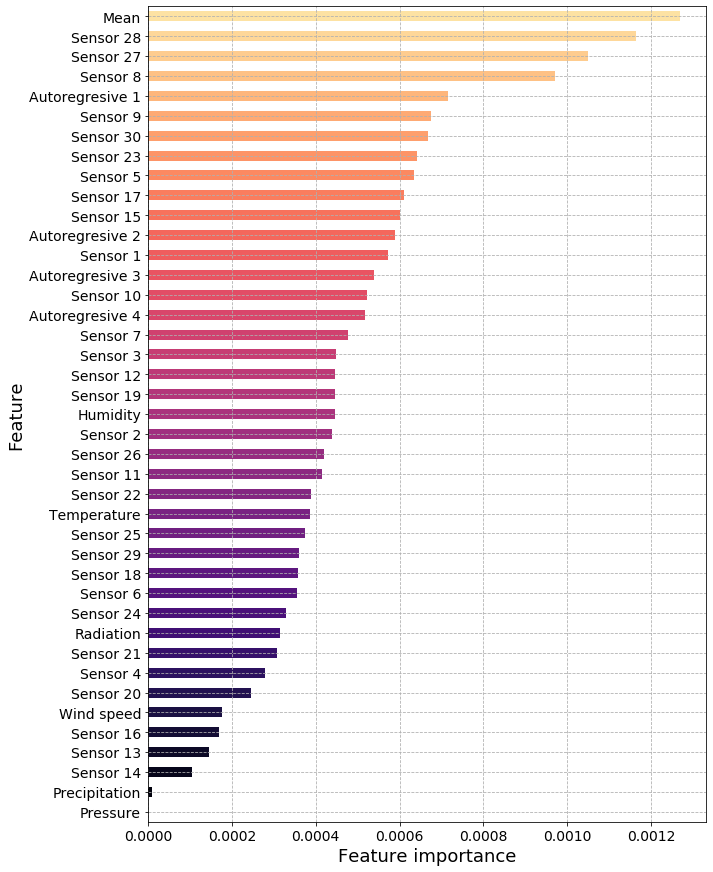}
  \caption{Mean SHAP values for all features in dense module computed in zone D. Temporal module
    output (\textit{Mean}), spatial module output (\textit{Sensors}),
    autoregresive terms and exogenous variables (both represented by their
    names).}
\label{fig:inter3}
\end{figure}

Finally, from average SHAP values computed for the dense module at Zone D (see
Sect. \ref{S3.2.4}), shown in Fig. \ref{fig:inter3}, we can extract several
conclusions. First of all, it supports the idea that using average traffic
intensity (``Mean'') for trend and seasonality modelling (temporal module) might
be beneficial. Secondly, the importance given to traffic sensors follow a
similar pattern to the one seen previously by the spatio-temporal attention
mechanism, reinforcing the idea of which spatial points are more important.
Thirdly, the autoregresive term that tries to capture the intertia of the series
seems to contribute positively too. Lastly, exogenous data importance points out
that it has the ability of improving the prediction significantly and should be
chosen carefully for each problem.

\section{Conclusions and future directions}
\label{S6}

Through this paper, a new spatio-temporal framework based on attention
mechanisms whose operation rest on several spatio-temporal series components is
presented. Unlike previous methodologies, we focus our efforts in creating a
system that can be considered robust and adaptable, evaluating it in a non-fixed
scenario. After being applied to a real traffic dataset, it has been proved that
outperforms four state of the art neural architectures and it has been studied
its behaviour respect to both time and spatial dimension through extensive
experimentation. By analyzing four different locations with 30 traffic sensors
each, we can confirm the statistical significance of our results with a
confidence of $95\%$ for forecasting horizons of up to 24 hours.

Thanks to the interpretable nature of the model, we have illustrated how that
information might be used in order to understand better how the framework works,
how it can give us specific information from the problem domain and why our
network architecture is well founded. Concretely, the conducted experiments have
shown that, as we postulated, the temporal module regulates seasonality and
trend, spatial module is capable of extracting short-term and spatial relations,
and that it is necessary to introduce explicit autoregressive terms in order to
exploit inertia correctly. Finally, these experiments demonstrate the
effectiveness of all these terms to make the final prediction.

For future work, it might be interesting to evaluate the proposed method over a
wider range of series in order to generalize the results and see its behaviour
over different applications. With the actual ability from the spatial module to
model attention for both input dimensions, space and time, it could be
beneficial to extend these idea to outputs dimensions too, having different
attention weights for different predicted timesteps. Lastly, it should be
studied how to use exogenous spatio-dependent data in the best possible way.

\section{Acknowledgements}
\label{S7}

This research has been partially funded by the \textit{Empresa Municipal de
  Transportes} (EMT) of Madrid under the program \textit{"Aula Universitaria
  EMT/UNED de Calidad del Aire y Movilidad Sostenible"}.

\bibliographystyle{./bibliography/IEEEtran}
\bibliography{./bibliography/bibliography.bib}

\end{document}